\documentclass{article}
\usepackage[english]{babel} 
\usepackage{times}
\usepackage{graphicx} 
\usepackage{subfigure} 
\usepackage{natbib}
\usepackage{hyperref}

\usepackage{algorithm}
\usepackage{algorithmicx}
\usepackage[noend]{algpseudocode}
\errorcontextlines\maxdimen

\makeatletter
    \newcommand*{\algrule}[1][\algorithmicindent]{\makebox[#1][l]{\hspace*{.5em}\thealgruleextra\vrule height \thealgruleheight depth \thealgruledepth}}%
\newcommand*{\thealgruleextra}{}
\newcommand*{\thealgruleheight}{.75\baselineskip}
\newcommand*{\thealgruledepth}{.25\baselineskip}

\newcount\ALG@printindent@tempcnta
\def\ALG@printindent{%
    \ifnum \theALG@nested>0
        \ifx\ALG@text\ALG@x@notext
        \else
            \unskip
            \addvspace{-1pt}
            \ALG@printindent@tempcnta=1
            \loop
                \algrule[\csname ALG@ind@\the\ALG@printindent@tempcnta\endcsname]%
                \advance \ALG@printindent@tempcnta 1
            \ifnum \ALG@printindent@tempcnta<\numexpr\theALG@nested+1\relax
            \repeat
        \fi
    \fi
    }%
\usepackage{etoolbox}
\patchcmd{\ALG@doentity}{\noindent\hskip\ALG@tlm}{\ALG@printindent}{}{\errmessage{failed to patch}}
\makeatother

\newbox\statebox
\newcommand{\myState}[1]{%
    \setbox\statebox=\vbox{#1}%
    \edef\thealgruleheight{\dimexpr \the\ht\statebox+1pt\relax}%
    \edef\thealgruledepth{\dimexpr \the\dp\statebox+1pt\relax}%
    \ifdim\thealgruleheight<.75\baselineskip
        \def\thealgruleheight{\dimexpr .75\baselineskip+1pt\relax}%
    \fi
    \ifdim\thealgruledepth<.25\baselineskip
        \def\thealgruledepth{\dimexpr .25\baselineskip+1pt\relax}%
    \fi
    \State #1%
    \def\thealgruleheight{\dimexpr .75\baselineskip+1pt\relax}%
    \def\thealgruledepth{\dimexpr .25\baselineskip+1pt\relax}%
}

\usepackage[final]{nips_2016}

\usepackage[utf8]{inputenc} 
\usepackage[T1]{fontenc}    
\usepackage{hyperref}       
\usepackage{url}            
\usepackage{booktabs}       
\usepackage{amsfonts}       
\usepackage{nicefrac}       
\usepackage{microtype}      

\usepackage{amsmath}
\usepackage{graphicx}
\usepackage[colorinlistoftodos]{todonotes}
\usepackage{caption}
\usepackage{pifont}
\usepackage{placeins}
\usepackage{mathtools}
\usepackage{natbib}
\usepackage{float}
\usepackage{placeins}
\usepackage{mathtools}
\usepackage{enumitem}
\usepackage{hyperref}

\usepackage{lipsum}

\title{Deep Spiking Networks}

\author{
  Peter O'Connor \\
  QUVA Lab\\
  University of Amsterdam\\
  Science Park 904, 1098 XH Amsterdam \\
  \texttt{p.e.oconnor@uva.nl} \\
  \And
    Max Welling\\
  QUVA Lab\\
  University of Amsterdam\\
  Science Park 904, 1098 XH Amsterdam \\
  \texttt{m.welling@uva.nl} \\
}

\begin{document}

\maketitle







\begin{abstract}
We introduce an algorithm to do backpropagation on a spiking network.  Our network is "spiking" in the sense that our neurons accumulate their activation into a potential over time, and only send out a signal (a ``spike'') when this potential crosses a threshold and the neuron is reset.  Neurons only update their states when receiving signals from other neurons.  Total computation of the network thus scales with the number of spikes caused by an input rather than network size.  We show that the spiking Multi-Layer Perceptron behaves identically, during both prediction and training, to a conventional deep network of rectified-linear units, in the limiting case where we run the spiking network for a long time.  We apply this architecture to a conventional classification problem (MNIST) and achieve performance very close to that of a conventional Multi-Layer Perceptron with the same architecture.  Our network is a natural architecture for learning based on streaming event-based data, and is a stepping stone towards using spiking neural networks to learn efficiently on streaming data.

\end{abstract}
\section{Introduction}
\label{sec:intro}

In recent years the success of Deep Learning has proven that a lot of problems in machine-learning can be successfully attacked by applying backpropagation to learn multiple layers of representation.  Most of the recent breakthroughs have been achieved through purely supervised learning.

In the standard application of a deep network to a supervised-learning task, we feed some input vector through multiple hidden layers to produce a prediction, which is in turn compared to some target value to find a scalar cost.  Parameters of the network are then updated in proportion to their derivatives with respect to that cost.  This approach requires that all modules within the network be differentiable.  If they are not, no gradient can flow through them, and backpropagation will not work.  

An alternative class of artificial neural networks are Spiking Neural Networks.  These networks, inspired by biology, consist of neurons that have some persistent ``potential'' which we refer to as $\phi$, and alter each-others' potentials by sending ``spikes'' to one another.  When unit $i$ sends a spike,  it increments the potential of each downstream unit $j$ in proportion to the synaptic weight $W_{i,j}$ connecting the units.  If this increment brings unit $j$'s potential past some threshold, unit $j$ sends a spike to its downstream units, triggering the same computation in the next layer.  Such systems therefore have the interesting property that the amount of computation done depends on the contents of the data, since a neuron may be tuned to produce more spikes in response to some pattern of inputs than another.

In our flavour of spiking networks, a single forward-pass is decomposed into a series of small computations provide successively closer approximations to the true output.  This is a useful feature for real time, low-latency applications, as in robotics, where we may want to act on data quickly, before it is fully processed.  If an input spike, on average, causes one spike in each downstream layer of the network, the average number of additions required per input-spike will be $\mathcal{O}(\sum_{l=1}^L N_l)$, where $N_l$ is the number of units in the layer $l$.  Compare this to a standard network, where the basic messaging entity is a vector.  When a vector arrives at the input, full forward pass will require $\mathcal{O}(\sum_{l=1}^L(N_{l-1}\cdot N_l))$ multiply-adds, and will yield no  ``preview'' of the network output. 

Spiking networks are well-adapted to handle data from event-based sensors, such as the Dynamic Vision Sensor (a.k.a. Silicon Retina, a vision sensor) \cite{lichtsteiner2008128} and the Silicon Cochlea (an audio sensor) \cite{chan2007aer}.  Instead of sending out samples at a regular rate, as most sensors do, these sensors asynchronously output events when there is a change in the input.  They can thus react with very low latency to sensory events, and produce very sparse data.  These events could be directly fed into our spiking network (whereas they would have to be binned over time and turned into a vector to be used with a conventional deep network).  

In this paper, we formulate a deep spiking network whose function is equivalent to a deep network of Rectified Linear (ReLU) units.  We then introduce a spiking version of backpropagation to train this network.    Compared to a traditional deep network, our Deep Spiking Network has the following advantageous properties:

\begin{enumerate}
\item Early Guessing.  Our network can make an ``early guess'' about the class associated with a stream of input events, before all the data has been presented to the network.
\item No multiplications.  Our training procedure consists only of addition, comparison, and indexing operations, which potentially makes it very amenable to efficient hardware implementation.  
\item Data-dependent computation.  The amount of computation that our network does is a function of the data, rather than the network size.  This is especially useful given that our network tends to learn sparse representations.
\end{enumerate}

The remainder of this paper is structured as follows:  In Section \ref{sec:relwork} we discuss past work in combining spiking neural networks and deep learning.  In \ref{sec:methods} we describe a Spiking Multi-Layer Perceptron.  In \ref{sec:experiment} we show experimental results demonstrating that our network behaves similarly to a conventional deep network in a classification setting.  In \ref{sec:discussion} we discuss the implications of this research and our next steps.

\section{Related Work}
\label{sec:relwork}

There has been little work on combining the fields of Deep Learning and Spiking neural networks.  The main reason for this is that there is not an obvious way to backpropagate an error signal through a spiking network, since output is a stream of discrete events, rather than smoothly differentiable functions of the input.  \cite{bohte2000spikeprop} proposes a spiking deep learning algorithm - but it involves simulating a dynamical system, is specific to learning temporal spike patterns, and has not yet been applied at any scale.  \cite{buesing2011neural} shows how a somewhat biologically plausible spiking network can be interpreted as an MCMC sampler of a high-dimensional probability distribution.  \cite{diehlfast} does classification on MNIST with a deep event-based network, but training is done with a regular deep network which is then converted to the spiking domain. A similar approach was used by \cite{hunsberger2015spiking} - they came up with a continuous unit which smoothly approximated the the firing rate of a spiking neuron, and did backpropagation on that, then transferred the learned parameters to a spiking network.  \cite{neftci2013event} came up with an event-based version of the contrastive-divergence algorithm, which can be used to train a Restricted Boltzmann Machine, but it was never applied in a Deep-Belief Net to learn multiple layers of representation.   \cite{o2013real} did create an event-based spiking Deep Belief Net and fed it inputs from event-based sensors, but the network was trained offline in a vector-based system before being converted to run as a spiking network.  

Spiking isn't the only form of discretization.  \cite{DBLP:journals/corr/CourbariauxBD15} achieved impressive results by devising a scheme for sending back an approximate error gradient in a deep neural network using only low-precision (discrete) values, and additionally found that the discretization served as a good regularizer.  Our approach (and spiking approaches in general) differ from this in that they sequentially compute the inputs over time, so that it is not necessary to have finished processing all the information in a given input to make a prediction.

\section{Methods}
\label{sec:methods}

In Sections \ref{sec:varspikequant} to\ref{sec:quantrec} we describe the components used in our model.  In Section \ref{sec:construct-smlp} we will use these components to put together a Spiking Multi-Layer Perceptron.

\subsection{Spiking Vector Quantization}
\label{sec:varspikequant}

\begin{figure}
\centering

\begin{minipage}{.45\textwidth}

  \begin{algorithm}[H]
    \captionsetup{width=0.95\textwidth}
    \caption{Spiking Vector Quantization }
    \label{alg:spiking-vector-quantization}
    \begin{algorithmic}[1]
      \State {\bfseries Input:} $\vec{v}\in\mathbb{R}^d$, $T\in\mathbb{N}$
      \State {\bfseries Internal:} $\vec{\phi}\in\mathbb{R}^d \leftarrow \vec{0}$ 
      \For{t $\in$ $1...T$ } 
      \State $\vec{\phi} \leftarrow \vec{\phi} + \vec{v}$
      \While{$max(|\vec{\phi}|)>\tfrac12$}
        \State $i \leftarrow argmax(|\vec{\phi}|)$
        \State $s \leftarrow sign(\phi_i)$
        \State $\vec{\phi}_i \leftarrow \vec{\phi}_i -s$
        \State FireSpike(source = i, sign = s)
      \EndWhile
      \EndFor
    \end{algorithmic}
    \captionsetup{width=0.95\textwidth}
  \end{algorithm}
\end{minipage}
\begin{minipage}{.45\textwidth}
  \centering
  \captionsetup{width=0.9\textwidth}
 \includegraphics[width=01\textwidth]{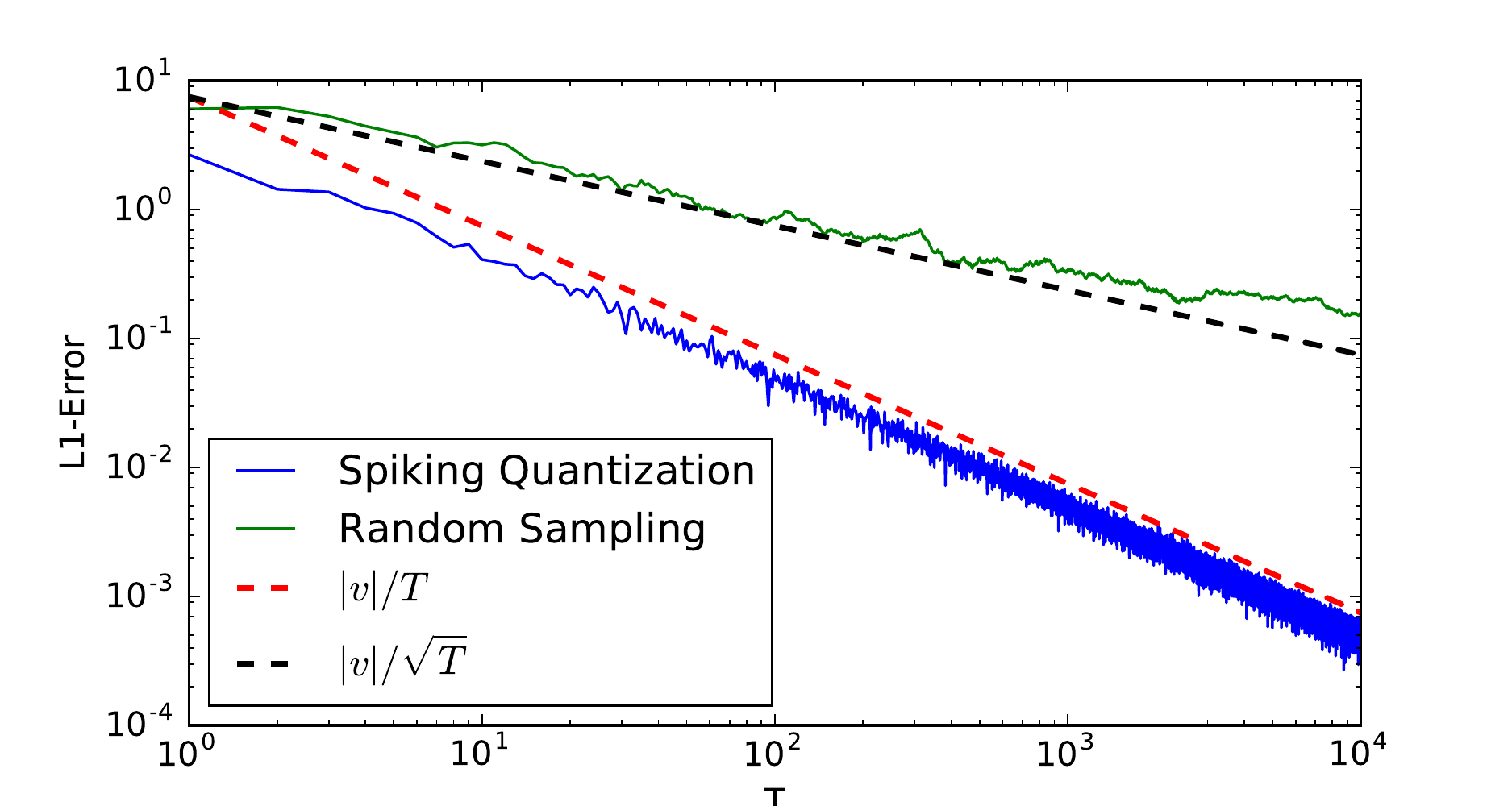}
\caption{\label{fig:convergence} Variable-Spike Quantization shows $1/T$ convergence, while ordinary sampling converges at a rate of $1/\sqrt{T}$.  Note both x and y axes are log-scaled.}
\end{minipage}%
\end{figure}

The neurons in the input layer of our network use an algorithm that we refer to as Spiking Vector Quantization (Algorithm \ref{alg:spiking-vector-quantization}) to generate ``signed spikes'' - that is, spikes with an associated positive or negative value.  Given a real vector: $\vec{v}$, representing the input to an array of neurons, and some number of time-steps $T$, the algorithm generates a series of $N$ signed-spikes: $\left<(i_n, s_n): i_n\in[1..len(\vec{v})], s_n\in\{\pm 1\}, n\in[1..N]\right>$, where  $N$ is the total number of spikes generated from running for $T$ steps, $i_n$ is the index of the neuron from which the $n$'th spike fires (note that zero or more spikes can fire from a neuron within one time step), $s_n\in \{\pm 1\}$ is the sign of the $n$'th spike.  

 In Algorithm \ref{alg:spiking-vector-quantization}, we maintain an internal vector of ``neuron potentials'' $\vec{\phi}$.  Every time we emit a spike from neuron $i$ we subtract $s_i$ from the potential  $\phi_i$ until $\vec{\phi}$ is in the interval bounded by $(-\frac12,\frac12)^{len(\vec{v})}$.  We can show that as we run the algorithm for a longer time (as $T\rightarrow\infty$), we observe the following limit:

\begin{equation}
\label{eq:spiking-vector-approx}
\lim_{T\to\infty}: \vec{v} = \frac1T \sum_{n=1}^{N} \vec{e_{i_n}} s_n 
\end{equation}
Where $\vec{e_{i_n}}$ is an one-hot encoded vector with index $i_n$ set to 1.  The proof is in the supplementary material.

Our algorithm is simply doing a discrete-time, bidirectional version of Delta-Sigma modulation - in which we encode floating point elements of our vector $\vec{v}$ as a stream of signed events.  We can see this as doing a sort of ``deterministic sampling'' or ``herding'' \cite{welling2009herding} of the vector v. Figure \ref{fig:convergence} shows how the cumulative vector from our stream of events approaches the true value of v at a rate of $1/T$.  We can compare this to another approach in which we stochastically sample spikes from the vector $\vec{v}$ with probabilities proportional to the magnitude of elements of $\vec v$, (see the ``Stochastic Sampling'' section of the supplementary material), which has a convergence of $1/\sqrt{T}$.

\subsection{Spiking Stream Quantization}

A small modification to the above method allows us to turn a stream of vectors into a stream of signed-spikes.

If instead of a fixed vector $\vec{v}$ we take a stream of vectors $v_{stream} = \{\vec{v_1}, ...\vec{v_T}\}$, we can modify the quantization algorithm to increment $\vec{\phi}$ by $\vec{v_t}$ on timestep $t$.  This modifies Equation \ref{eq:spiking-vector-approx} to:

\begin{equation} 
\lim_{T\to\infty}:  \frac1T\sum_{t=1}^T \vec{v_t}=\frac1T\sum_{n=1}^N\vec{e_{i_n}}s_n
\end{equation}
So we end up approximating the running mean of $v_{stream}$.  See ``Spiking Stream Quantization'' in the supplementary material for full algorithm and explanation.  When we apply this to implement a neural network in Section \ref{sec:construct-smlp}, this stream of vectors will be the rows of the weight matrix indexed by the incoming spikes.


\subsection{Rectifying Spiking Stream Quantization}
\label{sec:quantrec}

We can add a slight tweak to our Spiking Stream Quantization algorithm to create a spiking version of a rectified-linear (ReLU) unit.  To do this, we only fire events on positive threshold-crossings, resulting in Algorithm \ref{alg:rect-quantization-stream}.

  \begin{algorithm}[H]
    \captionsetup{width=0.95\textwidth}
    \caption{Rectified Spiking Stream Quantization }
    \label{alg:rect-quantization-stream}
    \begin{algorithmic}[1]
      \State {\bfseries Input:} $\vec{v_t}\in\mathbb{R}^d$, $t\in[1..T]$
      \State {\bfseries Internal:} $\vec{\phi}\in\mathbb{R}^d \leftarrow \vec{0}$ 
      \For{t $\in$ $1...T$ } 
      \State $\vec{\phi} \leftarrow \vec{\phi} + \vec{v_t}$
      \While{$max(\vec{\phi})>\tfrac12$}
        \State $i \leftarrow argmax(\vec{\phi})$
        \State $\vec{\phi}_i \leftarrow \vec{\phi}_i - 1$
        \State FireSpike(source = i, sign = +1)
      \EndWhile
      \EndFor
    \end{algorithmic}
    \captionsetup{width=0.95\textwidth}
  \end{algorithm}
  
 We can show that if we draw spikes from a stream of vectors in the manner described in Algorithm \ref{alg:rect-quantization-stream}, and sum up our spikes, we approach the behaviour of a ReLU layer: 
 
\begin{equation} 
\label{eq:rect-quant}
\lim_{T\to\infty}:  \max\left(0, \frac1T\sum_{t=1}^T \vec{v_t}\right)=\frac1T\sum_{n=1}^N\vec{e_{i_n}}
\end{equation}

 See ``Rectified Stream Quantization'' in the supplementary material for a more detailed explanation.

\subsection{Incremental Dot-Product}

Thus far, we've shown that our quantization method transforms a vector into a stream of events.  Here we will show that this can be used to incrementally approximate the dot product of a vector and a matrix.  Suppose we define a vector $\vec{u} \leftarrow \vec{v} \cdot W$, Where W is a matrix of parameters.  Given a vector $\vec{v}$, and using Equation \ref{eq:spiking-vector-approx}, we see that we can approximate the dot product with a sequence of additions:
\begin{equation}
\label{eq:inc-dot}
\begin{aligned}
&\vec{u} = \vec{v} \cdot W
\approx\frac1T\big(\sum_{n=1}^N\vec{e}_{i_n}s_n\big)\cdot W 
=\frac1T\big(\sum_{n=1}^N s_n \vec{e}_{i_n}\cdot W\big) 
=\frac1T\big(\sum_{n=1}^N s_n \vec{W}_{i_n,\cdot}\big) 
\end{aligned}
\end{equation}
Where $W_{i, \cdot}$ is the $i$'th row of matrix W.  

\subsection{Forward Pass of a Neural Network}
\label{sec:construct-smlp}

Using the parts we've described so far, Algorithm \ref{alg:forward-pass} describes the forward pass of a neural network.   The InputLayer procedure demonstrates how Spike Vector Quantization, shown in Algorithm \ref{alg:spiking-vector-quantization} transforms the vector into a stream of events.  The HiddenLayer procedure shows how we can combine the Incremental Dot-Product (Equation \ref{eq:inc-dot}) and Rectifying Spiking Stream Quantization (Equation \ref{eq:rect-quant}) to approximate the a fully-connected ReLU layer of a neural network.  The Figure in the "MLP Convergence" section of the supplimentary material shows that our spiking network, if run for a long time, exactly approaches the function of the ReLU network.

\begin{algorithm}
  \captionsetup{width=0.95\textwidth}
  \caption{Pseudocode for a forward pass in a network with one hidden layer}
  \label{alg:forward-pass}
  \begin{algorithmic}[1]
  \let\oldReturn\Return
  \renewcommand{\Return}{\State\oldReturn}

  \Function{ForwardPass}{$\vec{x}\in\mathbb{R}^{d_{in}}$, $T\in\mathbb{N}$}
    \State {\bfseries Variable:} $\vec{u} \in \mathbb{R}^{d_{out}}\leftarrow \vec{0}$

    \For{t $\in$ $1...T$ } 
      \State InputLayer($\vec{x}$) 
    \EndFor
    \Return $\vec{u}/T$

    \Procedure{InputLayer}{$\vec{v}\in\mathbb{R}^{d_{in}}$}
        \State {\bfseries Internal:} $\vec{\phi}\in \mathbb{R}^{d_{in}}$
        \State $\vec{\phi}\Leftarrow\vec{\phi}+\vec{v}$
        \While{$max(|\vec{\phi}|) > \tfrac12$}
            \State $i \leftarrow argmax(|\vec{\phi}|)$
            \State s = $sign(\phi_i)$ 
            \State $\phi_i\leftarrow\phi_i-s$
            \State HiddenLayer(i, s)
        \EndWhile
    \EndProcedure

    \Procedure{HiddenLayer}{$i \in [1..d_{in}]$, $s\in[-1, +1]$}
        \State {\bfseries Internal:} $\vec{\phi}\in \mathbb{R}^{d_{hid}}$, $W\in\mathbb{R}^{d_{in} \times d_{hid}}$
        \State $\vec{\phi}\Leftarrow\vec{\phi}+s\cdot W_{i,\cdot}$
        \While{$max(\vec{\phi}) > \tfrac12$}
            \State $i \leftarrow argmax(\vec{\phi})$
            \State $\phi_i\leftarrow\phi_i-1$
            \State OutputLayer(i)
        \EndWhile
    \EndProcedure

    \Procedure{OutputLayer}{$i \in [1..d_{hid}]$}
        \State {\bfseries Internal:} $W\in\mathbb{R}^{d_{hid} \times d_{out}}$
        \State {\bfseries Global:} $\vec{u}\leftarrow \vec{u}+W_{i,\cdot}$  \Comment changes $\vec{u}$
    \EndProcedure

  \EndFunction
  \end{algorithmic}
  \captionsetup{width=0.95\textwidth}
\end{algorithm}

\subsection{Backward Pass}
\label{sec:backpass}

In the backwards pass we propagate error spikes backwards, in the same manner as we propagated the signal forwards, so that the error spikes approach the true gradients of the ReLU network as $T\rightarrow\infty$.  Pseudocode explaining the procedure is provided in the ``Training Iteration'' Section of the supplementary material, and a diagram explaining the flow of signals is in the ``Network Diagram'' section.  

A ReLU unit has the function and derivative:

\begin{equation}
\begin{aligned}
f(x) &= [x>0]\cdot x \\
f'(x)&=[x>0]
\end{aligned}
\end{equation}
Where: \\
$[x>0]$  denotes a step function (1 if $x>0$ otherwise 0).

In the spiking domain, we express this simply by blocking error spikes on units for which the cumulative sum of inputs into that unit is below 0 (see the "filter" modules in the ``Network Diagram'' section of the supplementary material).

The signed-spikes that represent the backpropagating error gradient at a given layer are used to index columns of that layer's weight matrix, and negate them if the sign of the spike is negative.  The resulting vector is then quantized, and the resulting spikes are sent back to previous layers.  

One problem with the scheme described so far is that, when errors are small, it is possible that the error-quantizing neurons never accumulate enough potential to send a spike before the training iteration is over.  If this is the case, we will never be able to learn when error gradients are sufficiently small.  Indeed, when initial weights are too low, and therefore the initial magnitude of the backpropagated error signal is too small, the network does not learn at all.  This is not a problem in traditional deep networks, because no matter how small the magnitude, some error signal will always get through (unless all hidden units are in their inactive regime) and the network will learn to increase the size of its weights.  We found that a surprisingly effective solution to this problem is to simply not reset the $\phi$ of our error quantizers between training iterations.  This way, after some burn-in period, the quantizer's $\phi$ starts each new training iteration at some random point in the interval $[-\frac12, \frac12]$, and the unit always has a chance to spike.

A further issue that comes up when designing the backward pass is the order in which we process events.  Since an event can move a ReLU unit out of its active range, which blocks the transmission of itself or future events on the backward pass, we need to think about the order in which we processing these events.  The topic of event-routing is explained in the ``Event Routing'' section of the supplementary material.

\subsection{ Weight Updates}
\label{sec:weightupdates}

We can collect spike statistics and generates weight updates.  There are two methods by which we can update the weights.  These are as follows:

\textbf{Stochastic Gradient Descent}
The most obvious method of training is to approximate stochastic gradient descent. In this case, we accumulate two spike-count vectors, $\vec c_{in}$ and $\vec c_{error}$ and take their outer product at the end of a training iteration to compute the weight update:
\begin{equation}
\Delta W \leftarrow -\frac{\eta}T \cdot \vec{c_{in}} \otimes  \vec c_{error} \approx -\frac{\eta}T \frac{\partial \mathcal{L}}{\partial W}
\end{equation}

\textbf{Fractional Stochastic Gradient Descent (FSGD)}
We can also try some thing new.  Our spiking network introduces a new feature: if a data point is decomposed as a stream of events, we can do parameter updates even before a single data point has been observed.  If we do updates whenever an error event comes back, we update each weight based on only the input data that has been seen \textit{so far}.  This is described by the rule:

\begin{equation}
\Delta W_{:,i} \leftarrow -\frac{\eta}T \cdot s \cdot \vec{c_{in}}
\end{equation}
Where $\vec{c_{in}}$ is an integer vector of counted input spikes, $\Delta W_{:,i}$ is the change to the $i'th$ column of the weight matrix, $s \in \{-1, 1\}$ is the sign of the error event, and $i$ is the index of the unit that produced that error event, and $T$ is the number of time-steps per training iteration.  Early input events will contribute to more weight updates than those seen near the end of a training iteration.  Experimentally (see Section \ref{sec:exp_relu_comparison}, we see that this works quite well.  It may be that the additional influence given to early inputs causes the network to learn to make better predictions earlier on, compensating for the approximation caused by finite-runtime of the network.

\subsection{Training}
We chose to train the network with one sample at a time, although in principle it is possible to do minibatch training.  We select a number of time steps $T$, to run the network for each iteration of training.  At the beginning of a round of training, we reset the state of the forward-neurons (all $\vec{\phi}$'s and the state of the running sum modules), and leave the state of the error-quantizing neurons (as described in \ref{sec:backpass}).  On each time step $t$, we feed the input vector to the input quantizer, and propagate the resulting spikes through the network.  We then propagate an error spike back from the unit corresponding to the correct class label, and update the parameters by one of the two methods described in \ref{sec:weightupdates}.  See the ``Network Diagram'' section of the supplementary material to get an idea of the layout of all the modules.

\FloatBarrier
\section{Experiments}
\label{sec:experiment}

\subsection{Simple Regression}

We first test our network as a simple regressor, (with no hidden layers) on a binarized version of the newsgroups-20 dataset, where we do a 2-way classification between the electronics and medical newsgroups based word-count vectors.  We split the dataset with a 7-1 training-test ratio (as in \cite{crammer2009adaptive}) but do not do cross-validation.  Table \ref{tab:newsgroups-scores} shows that it works.

\begin{table}[!htb]
\begin{center}
 \begin{tabular}{|l|c|} 
 \hline
\textbf{Network} & \% \textbf{Test / Training Error} \\ [0.5ex] 
 \hline\hline
1 Layer NN & 2.278 / 0.127\\ 
Spiking Regressor & 2.278 / 0.82\\ 
SVM & 4.82 / 0\\
 \hline
\end{tabular}
\end{center}
 \caption{Scores on 20 newsgroups, 2-way classification between 'med' and 'electronic' newsgroups.  We see that, somewhat surprisingly, our approach outperforms the SVM.  This is probably because, being trained through SGD and tested at the end of each epoch, our classifier had more recently learned on samples at the end of the training set, which are closer in distribution to the test set than those at the beginning.}
  \label{tab:newsgroups-scores}
\end{table}

\subsection{Comparison to ReLU Network on MNIST}
\label{sec:exp_relu_comparison}
We ran both the spiking network and the equivalent ReLU network on MNIST, using an architecture with 2 fully-connected hidden layers, each consisting of 300 units.  Refer to the ``Hyperparameters'' section of the Supplimentary Material for a full description of hyperparameters.  

\begin{table}[!htb]
\begin{center}
 \begin{tabular}{|l|c|} 
 \hline
\textbf{Network} & \% \textbf{Test / Training Error} \\ [0.5ex] 
 \hline\hline
Spiking SGD:  & 3.6 / 2.484 \\
Spiking FSGD: & 2.07 / 0.37\\
Vector ReLU MLP & 1.63 / 0.426 \\ 
Spiking with ReLU Weights & 1.66 / 0.426 \\
ReLU with Spiking FSGD weights & 2.03 / 0.34\\
 \hline
\end{tabular}
\end{center}
 \caption{Scores of various implementations on MNIST after 50 epochs of training on a network with hidden layers of sizes [300, 300].  ``Spiking SDG''  and ``Spiking FSGD'' are  the spiking network trained with Stochastic Gradient Descent and Fractional Stochastic Gradient descent, respectively, as described in Section \ref{sec:weightupdates}.  ``Vector ReLU MLP'' is the score of a conventional MLP with ReLU units and the same architecture and training scheme.   ``Spiking with ReLU Weights'' is the score if we set the parameters of the Spiking network to the already-trained parameters of the ReLU network, then use the Spiking Network to classify MNIST. ``ReLU with Spiking weights'' is the inverse - we take the parameters trained in the spiking FSGD network and map them to the ReLU net.}
  \label{tab:mnist-scores}
\end{table}

\begin{figure}[!htb]
\centering
\includegraphics[width=.65\textwidth]{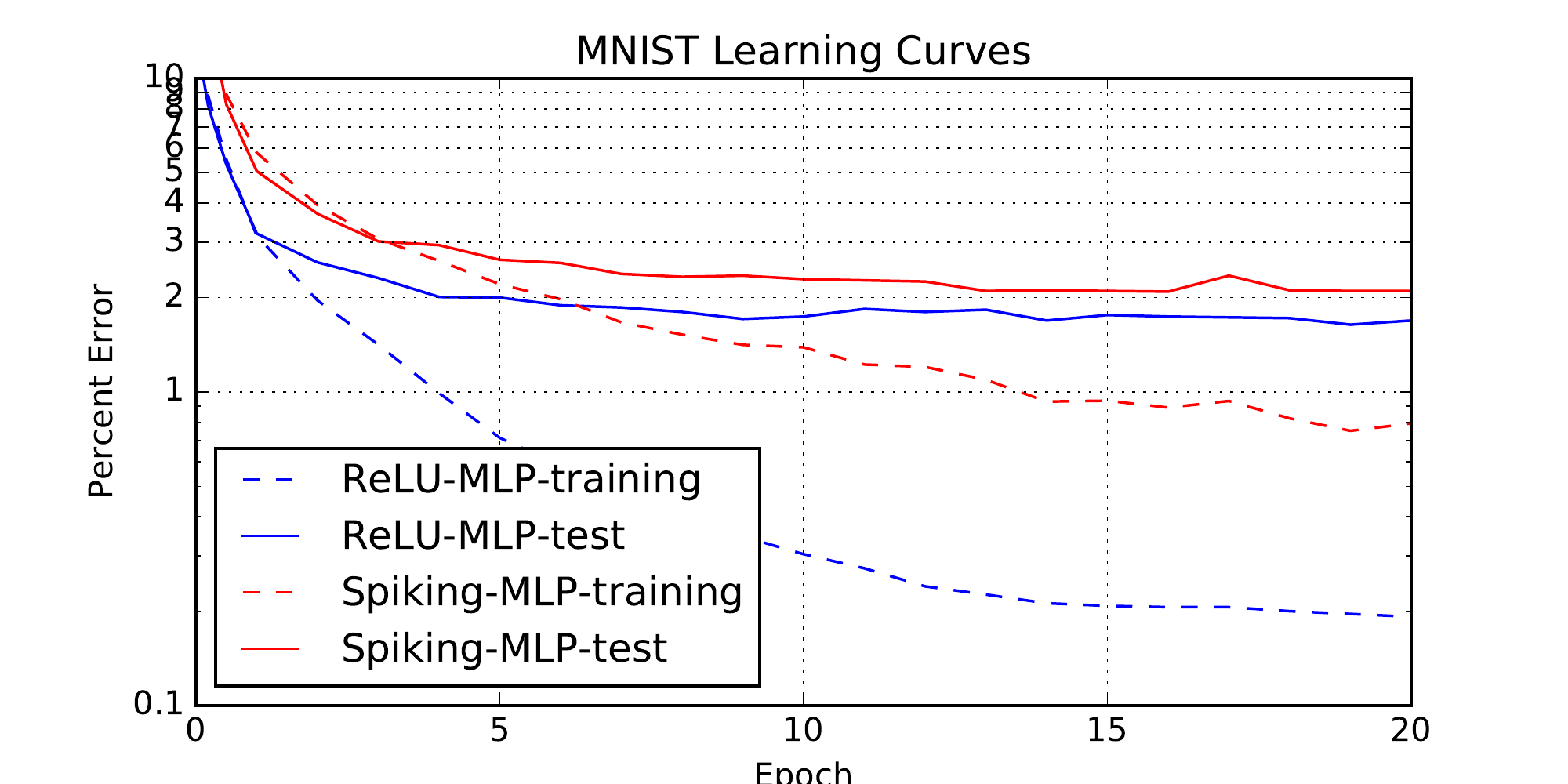}
\caption{\label{fig:mnist-training}  The Learning curves of the ReLU network (blue) and the Spiking network (red).  Solid lines indicate test error and dashed lines indicate training error.}
\end{figure}

Table \ref{tab:mnist-scores} shows the results of our experiment, after 50 epochs of training.  We find that the conventional ReLU network outperforms our spiking network, but only marginally.  In order to determine how much of that difference was due to the fact that the Spiking network has a discrete forward pass, we mapped the learned parameters from the ReLU network onto the Spiking network (spiking with ReLU Weights''), and used the Spiking network to classify .  The performance of the spiking network improved nearly to that of the ReLU network , indicating that the difference was not just due to the discretization of the forward pass but also due to the parameters learned in training.  We also did the inverse (ReLU with Spiking-FSGD-trained weights) - map the parameters of the trained Spiking Net onto the ReLU net, and found that the performance became very similar to that of the original Spiking (FSGD-Trained) Network.  This tells us that most of the difference in score is due to the approximations in training, rather than the forward pass.  Interestingly, our Spiking-FSGD approach outperforms the Spiking-SGD - it seems that by putting more emphasis on early events, we compensate for the finite runtime of the Spiking Network.  Figure \ref{fig:mnist-training} shows the learning curves over the first 20-epochs of training.  We see that the gap between training and test performance is much smaller in our Spiking network than in the ReLU network, and speculate that this may be to the regularization effect of the spiking.  To confirm this, we would have to show that on a larger network, our regularization actually helps to prevent overfitting.

\subsection{Early Guessing}

\begin{figure}[H]
\centering
\begin{minipage}{.45\textwidth}
  \centering
  \captionsetup{width=0.9\textwidth}
 \includegraphics[width=01\textwidth]{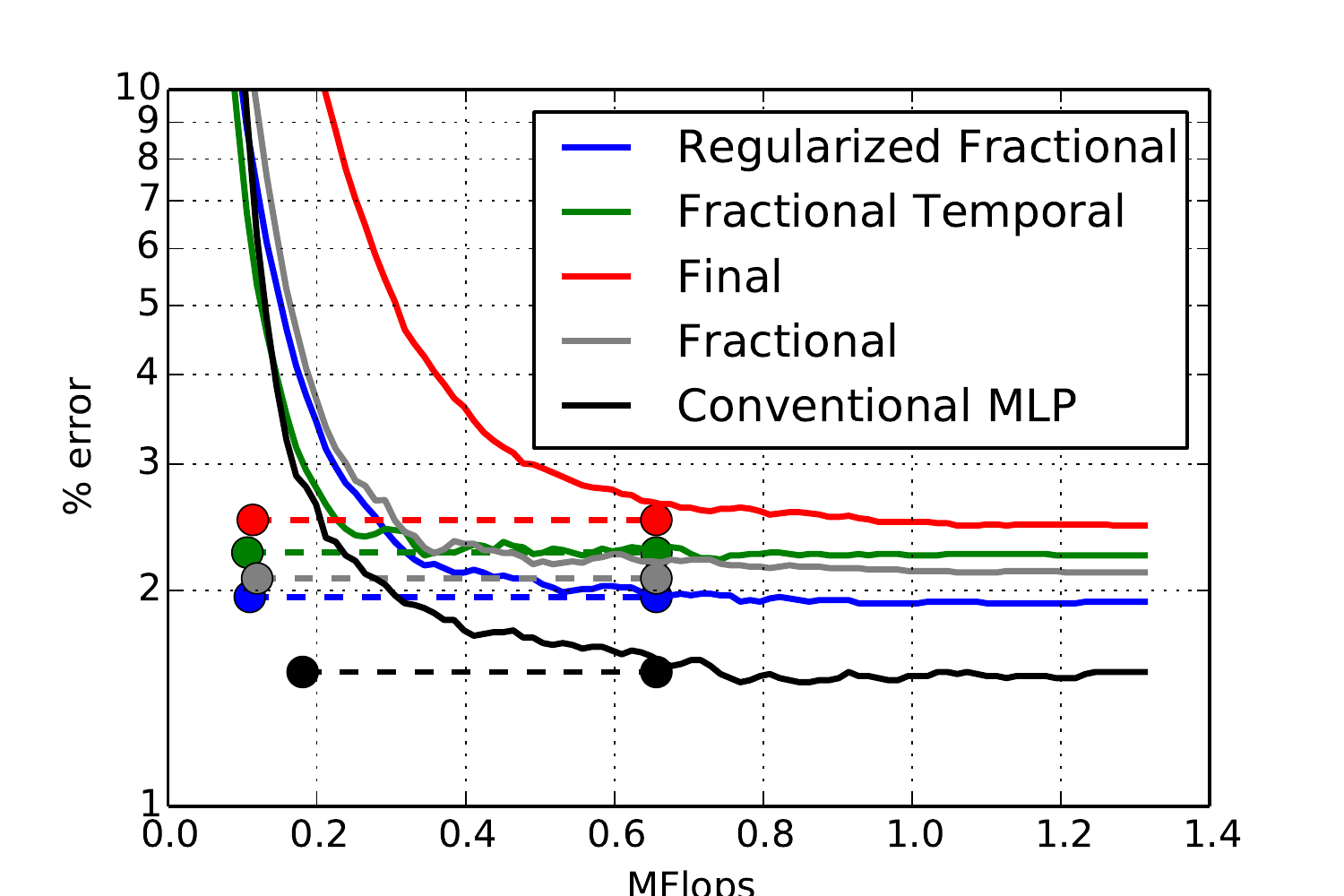}
\end{minipage}%
\begin{minipage}{.45\textwidth}
\centering
\includegraphics[width=1\textwidth]{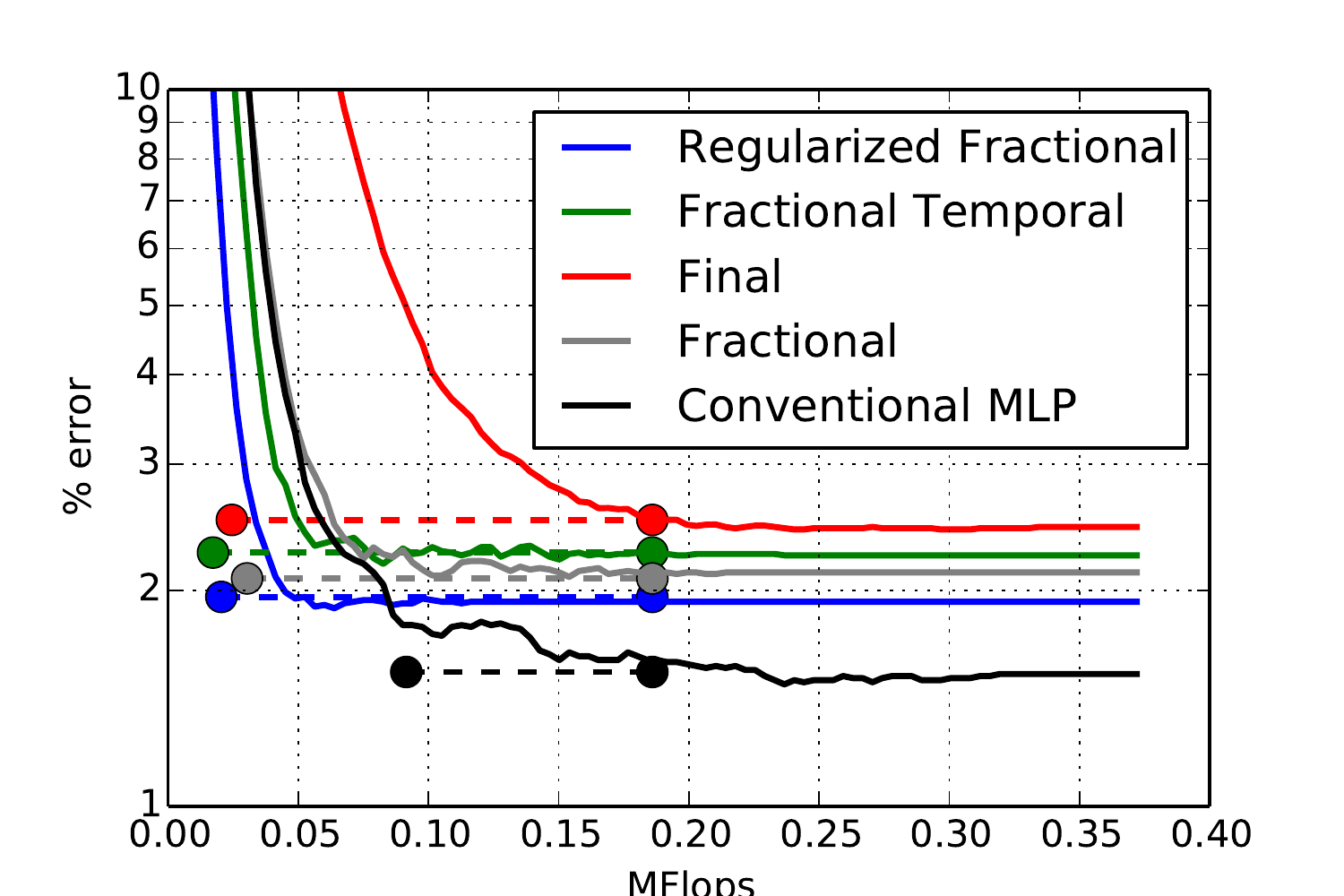}

\end{minipage}
\caption{\label{fig:flop-counts} Left: We compare the total computation (x-axis, in MegaFlops) required to achieve a given score (y-axis, in percent error), between differently trained predictors.  Each curve represents a differently trained spiking network (Regularized Fraction was trained with a regularization term, Fractional Temporal was trained with higher learning rate for early events).  The black line is the convergence curve of a spiking network with parameters learned in a conventional ReLU net of the same architecture (784-300-300-10).  The dots show the computation time and performance of a conventional ReLU network, with the right dot indicating the cost of a full feedforward pass, and the left indicating the cost when one removes units with 0-activation when computing the cost of a matrix multiplication.   We see that, when considering the full network, our spiking approach does give a good early guess compared to a naively implemented deep net, but not after considering sparsity.  Right: However, when only considering layers after the input layer (whose sparsity we do not control), we can see that there is an advantage to our spiking training scheme: In the low-flop range our spiking nets have lower error.  The networks that were trained as spiking networks are better at making early guesses than the conventionally trained network.  
}
\end{figure}

We evaluated the "early guess" hypothesis from Section \ref{sec:intro} using MNIST.  The hypothesis was that our spiking network should be able to make computational cheap ``early guesses" about the class of the input, before actually seeing all the data.  A related hypothesis was under the ``Fractional'' update scheme discussed in Section \ref{sec:weightupdates}, our networks should learn to make early guesses more effectively than networks trained under regular Stochastic Gradient Descent, because early input events contribute to more weight updates than later ones.  Figure \ref{fig:flop-counts} shows the results of this experiment.   We find, unfortunately, that our first hypothesis does not hold.  The early guesses we get with the spiking network cost more than a single (sparse) forward pass of the input vector would.  The second hypothesis, however, is supported by the right-side of Figure 5.  Our networks trained with Fractional Stochastic Gradient Descent make better early guesses than those trained on regular SGD.

\section{Discussion}
\label{sec:discussion}

We implemented a Spiking Multi-Layer Perceptron and showed that our network behaves very similarly to a conventional MLP with rectified-linear units.  However, our model has some advantages over a regular MLP, most of which have yet to be explored in full.  Our network needs neither multiplication nor floating-point numbers to work.  If we use Fractional Stochastic Gradient Descent, and scale all parameters in the network (initial weights, thresholds, and the learning rate) by the inverse of the learning rate, the only operations used are integer addition, indexing, and comparison.  This makes our system very amenable to efficient hardware implementation.

The Spiking MLP brings us one step closer to making a connection between the types of neural networks we observe in biology and the type we use in deep learning.  Like biological neurons, our units maintain an internal potential, and only communicate when this potential crosses some firing threshold. We believe that the main value of this approach is that it is a stepping stone towards a new type of deep learning.   The way that deep learning is done now takes no advantage of the huge temporal redundancy in natural data.  In the future we would like to adapt the methods developed here to work with nonstationary data.   Such a network could pass spikes to keep the output distribution in ``sync'' with an ever-changing input distribution.  This property - efficiently keeping an online track of latent variables in the environment, could bring deep learning into the world of robotics.

\clearpage
\bibliographystyle{plainnat}
\bibliography{mybib}

\clearpage
\appendix

\section{Algorithms}
\label{app:algorithms}

\subsection{Proof of convergence of Spike-Vector Quantization}

Here we show that if we obtain events $\left<(i_n, s_n): n\in[1..N]\right>$ given a vector $\vec{v}$ and a time $T$ from the Spiking-Vector Quantization Algorithm then:

\begin{equation}
\label{eq:spiking-vector-approx}
\lim_{T\rightarrow\infty} \vec{v} = \frac1T \sum_{n=1}^{N} \vec{e_{i_n}} s_n 
\end{equation}

  \begin{algorithm}[H]
    \captionsetup{width=0.95\textwidth}
    \caption{Spiking Vector Quantization }
    \label{alg:quantization}
    \begin{algorithmic}[1]
      \State {\bfseries Input:} $\vec{v}\in\mathbb{R}^d$, $T\in\mathbb{N}$
      \State {\bfseries Internal:} $\vec{\phi}\in\mathbb{R}^d \leftarrow \vec{0}$ 
      \For{t $\in$ $1...T$ } 
      \State $\vec{\phi} \leftarrow \vec{\phi} + \vec{v}$
      \While{$max(|\vec{\phi}|)>\tfrac12$}
        \State $i \leftarrow argmax(|\vec{\phi}|)$
        \State $s \leftarrow sign(\phi_i)$
        \State $\vec{\phi}_i \leftarrow \vec{\phi}_i -s$
        \State FireSpike(source = i, sign = s)
      \EndWhile
      \EndFor
    \end{algorithmic}
    \captionsetup{width=0.95\textwidth}
  \end{algorithm}

Since  $\forall i: -\frac12<\phi_i<\frac12$ the L1 norm is bounded by: 
\begin{equation} 
\|\phi_{T}\|_{L1} = \|\sum_{t=1}^T v - \sum_{n=1}^N e_{i_n}s_n\|_{L1}  < \frac{l(\vec{v})}{2}
\end{equation} 
where $l(\vec{v})$ is the number of elements in vector $\vec{v}$.
We can take the limit of infinite time, and show that our spikes converge to form an approximation of $\vec{v}$:

\begin{equation} 
\label{eq:herd-conv}
\begin{aligned}
&\lim_{T\to\infty}:  \frac1T\| \phi_T \|_{L1} = 
\lim_{T\to\infty}:   \| \frac1T \sum_{t=1}^T v -\frac1T\sum_{n=1}^N\vec{e_{i_n}} s_n \|_{L1}=0 \\
&\lim_{T\to\infty}:  \vec{v}=\frac1T\sum_{n=1}^N\vec{e_{i_n}}s_n
\end{aligned}
\end{equation}

\subsection{Stochastic Sampling}

\begin{algorithm}[H]
\caption{Stochastic Sampling of events from a vector}
\label{alg:stochastic}
\begin{algorithmic}[1]
\State {\bfseries Input:} vector $v$, int $T$
\State $mag \leftarrow sum(abs(\vec{v}))$
\State $\vec{p} = abs(\vec{v})/mag$
\For{t \Pisymbol{psy}{206} $1...T$ } 
\State $N=poisson(mag)$
  \For{n \Pisymbol{psy}{206} $1...N$ } 
\State $i \leftarrow DrawSample(\vec{p})$
\State $s \leftarrow sign(v_i)$
\State FireSignedSpike(index = i, sign = s)
\EndFor
\EndFor
\end{algorithmic}
\end{algorithm}

\subsection{Spiking Stream Quantization}

In our modification to Spiking Vector Quantization, we instead feed in a stream of vectors, as in Algorithm \ref{alg:quantization-stream}.

  \begin{algorithm}[H]
    \captionsetup{width=0.95\textwidth}
    \caption{Spiking Stream Quantization }
    \label{alg:quantization-stream}
    \begin{algorithmic}[1]
      \State {\bfseries Input:} $\vec{v_t}\in\mathbb{R}^d$, $t\in[1..T]$
      \State {\bfseries Internal:} $\vec{\phi}\in\mathbb{R}^d \leftarrow \vec{0}$ 
      \For{t $\in$ $1...T$ } 
      \State $\vec{\phi} \leftarrow \vec{\phi} + \vec{v_t}$
      \While{$max(|\vec{\phi}|)>\tfrac12$}
        \State $i \leftarrow argmax(|\vec{\phi}|)$
        \State $s \leftarrow sign(\phi_i)$
        \State $\vec{\phi}_i \leftarrow \vec{\phi}_i -s$
        \State FireSpike(source = i, sign = s)
      \EndWhile
      \EndFor
    \end{algorithmic}
    \captionsetup{width=0.95\textwidth}
  \end{algorithm}

If we simply replace the term $\sum_{t=1}^T \vec{v}$ in Equation \ref{eq:herd-conv} with $\sum_{t=1}^T \vec{v_t}$, and follow the same reasoning, we find that we converge to the running mean of the vector-stream.

\begin{equation} 
\label{eq:stream-convergence}
\begin{aligned}
&\lim_{T\to\infty}:  \frac1T\| \phi_T \|_{L1} = 
\lim_{T\to\infty}:   \| \frac1T \sum_{t=1}^T v_t -\frac1T\sum_{n=1}^N\vec{e_{i_n}} s_n \|_{L1}=0 \\
&\lim_{T\to\infty}:  \frac1T \sum_{t=1}^T \vec{v}=\frac1T\sum_{n=1}^N\vec{e_{i_n}}s_n
\end{aligned}
\end{equation}

\subsection{Rectified Stream Quantization}

We can further make a small modification where we only send positive spikes (so our $\vec{\phi}$ can get unboundedly negative.

  \begin{algorithm}[H]
    \captionsetup{width=0.95\textwidth}
    \caption{Rectified Spiking Stream Quantization }
    \label{alg:rect-quantization-stream}
    \begin{algorithmic}[1]
      \State {\bfseries Input:} $\vec{v_t}\in\mathbb{R}^d$, $t\in[1..T]$
      \State {\bfseries Internal:} $\vec{\phi}\in\mathbb{R}^d \leftarrow \vec{0}$ 
      \For{t $\in$ $1...T$ } 
      \State $\vec{\phi} \leftarrow \vec{\phi} + \vec{v_t}$
      \While{$max(\vec{\phi})>\tfrac12$}
        \State $i \leftarrow argmax(\vec{\phi})$
        \State $\vec{\phi}_i \leftarrow \vec{\phi}_i - 1$
        \State FireSpike(source = i, sign = +1)
      \EndWhile
      \EndFor
    \end{algorithmic}
    \captionsetup{width=0.95\textwidth}
  \end{algorithm}

To see why this construction approximates a ReLU unit, first observe that the total number of spikes emitted can be computed by considering the total cumulative sum $\sum_{t=1}^{T} v_{j,t}$. More precisely:
\begin{equation} 
\label{eq:n-rect-spikes}
N_{j,T} = \max\big(0, \big\lfloor \max_{T' \in [1...T]}\big(\sum_{t=1}^{T'} v_{j,t} + \frac12 \big) \big\rfloor\big)
\end{equation}
where $N_{j,T}$ indicates the number of spikes emitted from unit $j$ by time $T$ and $\lfloor \cdot \rfloor$ indicates the integer floor of a real number.

Assume the $v_{j,t}$ are IID sampled from some process with mean $E[v_{j,t}]=\mu_j$ and finite standard deviation $\sigma_j$. Define $\zeta_{j,t}=v_{j,t}-\mu_j$ which has zero mean and the cumulative sum $\xi_{j,T}=\sum_{t=1}^{T} \zeta_{j,t}$ which is martingale. There are a number of concentration inequalities, such as the Bernstein concentration inequalities \cite{fan2012hoeffding} that bound the sum or the maximum of the sequence $\xi_{j,T}$ under various conditions. What is only important for us is the fact that in the limit  $T\rightarrow\infty$ the sums $\xi_{j,T}$ concentrate to a delta peak at zero in probability and that we can therefore conclude $\sum_{t=1}^{T} v_{j,t} \rightarrow T\mu_j$ from which we can also conclude that the maximum, and thus the number of spikes will grow in the same way. From this we finally conclude that $\frac{N_{j,T}}{T}\rightarrow \max(0,\mu)$, which is the ReLU non-linearity.  Thus the mean spiking rate approaches the ReLU function of the mean input.

\section{MLP Convergence}

\begin{figure*}[!htb]
\centering
\includegraphics[width=1\textwidth]{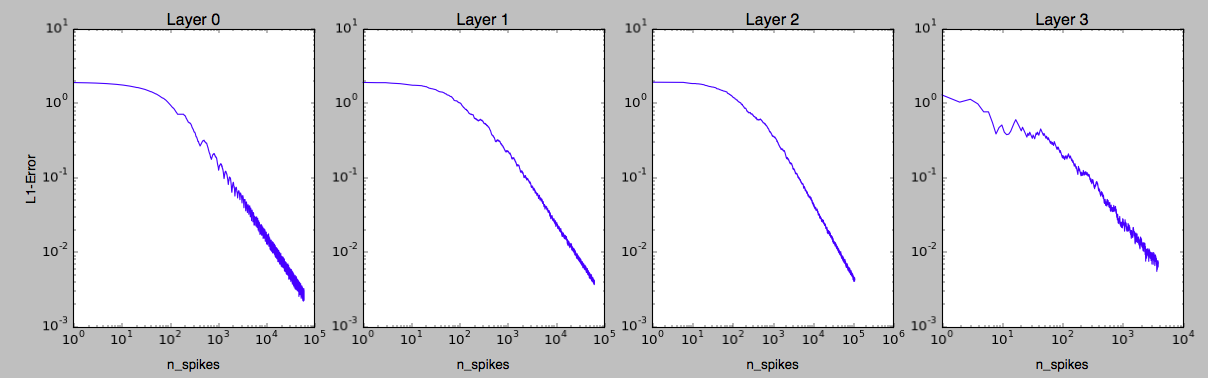}
\caption{\label{fig:mlp-convergence} A 3-layer MLP (784-500-500-10) MLP with random weights ($\sim \mathcal{N}(0, 0.1)$) is fed with a random input vector, and a forward pass is computed.  We compare the response of the ReLU network to the counts of spikes from our spiking network, and see that over all layers, the responses converge as $T\rightarrow\infty$.  Note both x and y axes are log-scaled. }
\end{figure*}

\section{A Training Iteration}

\begin{algorithm}[H]
  \captionsetup{width=0.95\textwidth}
  \caption{Pseudocode for a single Training Iteration on a network with one hidden layer with Fractional Stochastic Gradient Descent}
  \label{alg:forward-pass}
  \begin{algorithmic}[1]
  \let\oldReturn\Return
  \renewcommand{\Return}{\State\oldReturn}
  \algnewcommand{\Object}{\textbf{Object: }}
  \algdef{SE}[SUBALG]{Indent}{EndIndent}{}{\algorithmicend\ }%
  \algtext*{Indent}
  \algtext*{EndIndent}

  \Procedure{TrainingIteration}{$\vec{x}\in\mathbb{R}^{d_{in}}$, $\vec{y}\in\mathbb{R}^{d_{out}}$,  $T\in\mathbb{N}$, $\eta \in \mathbb{R}$}
    \State {\bfseries Variables:} $\vec{u} \in \mathbb{R}^{d_{out}}\leftarrow \vec{0}$, $W_{hid}\in\mathbb{R}^{d_{in}\times d_{hid}}$, $W_{out}\in\mathbb{R}^{d_{hid}\times d_{out}}$

    \For{t $\in$ $1...T$ } 
      \State InputLayer.forward($\vec{x}$) 
      \State OutputLayer.backward($\vec{u} - \vec{y}$)
    \EndFor

    \Procedure{InputLayer}{$\vec{v}\in\mathbb{R}^{d_{in}}$}
      \State {\bfseries Internal:} $\vec{\phi}\in \mathbb{R}^{d_{in}}$
      \State $\vec{\phi}\Leftarrow\vec{\phi}+\vec{v}$
      \While{$max(|\vec{\phi}|) > \tfrac12$}
        \State $i \leftarrow argmax(|\vec{\phi}|)$
        \State s = $sign(\phi_i)$ 
        \State $\phi_i\leftarrow\phi_i-s$
        \State HiddenLayer(i, s)
      \EndWhile
    \EndProcedure

	\Object{HiddenLayer}
    \Indent 
       \State {\bfseries Internal: } $\vec{c_{in}} \in \mathbb{R}^{d_{in}} \leftarrow \vec{0}$, $\vec{c_{out}} \in \mathbb{R}^{d_{hid}} \leftarrow \vec{0}$
       \Procedure{Forward}{$i \in [1..d_{in}]$, $s\in[-1, +1]$}
          \State {\bfseries Internal:} $\vec{\phi}\in \mathbb{R}^{d_{hid}}$
       	  \State $c_{in_i} \leftarrow c_{in_i} + s$
          \State $\vec{\phi}\Leftarrow\vec{\phi}+s\cdot [W_{hid}]_{i,\cdot}$
          \State $\vec{c_{out}} \leftarrow \vec{c_{out}}+[W_{hid}]_{i,\cdot}$
          \While{$max(\vec{\phi}) > \tfrac12$}
              \State $i \leftarrow argmax(\vec{\phi})$
              \State $\phi_i\leftarrow\phi_i-1$
              \State OutputLayer(i, +1)
          \EndWhile
      \EndProcedure
    
      \Procedure{Backward}{$\vec{v} \in \mathbb{R}^{d_{hid}}$}
      	  \State {\bfseries Internal:} $\vec{\phi} \in \mathbb{R}^{d_{hid}}$
          \State $\vec{\phi}\Leftarrow\vec{\phi}+\vec{v}\odot [\vec{c_{out}}>0]$
          \While{$max(|\vec{\phi}|) > \tfrac12$}
          \State $j \leftarrow argmax(|\vec{\phi}|)$
          \State s = $sign(\phi_i)$ 
          \State $\phi_j\leftarrow\phi_j-s$
          \State $[W_{hid}]_{\cdot, j} \leftarrow [W_{hid}]_{\cdot, j} - \eta \cdot s\cdot\vec{c_{in}}$ \Comment Update to $W_{hid}$
          \EndWhile
      \EndProcedure
	\EndIndent

	\Object{OutputLayer}
    \Indent
      \State {\bfseries Internal: } $\vec{c_{in}} \in \mathbb{R}^{d_{hid}} \leftarrow \vec{0}$
      \Procedure{Forward}{$i \in [1..d_{hid}]$, $s\in[-1, +1]$}
      	  \State $c_{in_i} \leftarrow c_{in_i} + s$
          \State {\bfseries Global:} $\vec{u}\leftarrow \vec{u}+s\cdot [W_{out}]_{i,\cdot}$  \Comment Update to $\vec{u}$
      \EndProcedure
      
      \Procedure{Backward}{$v \in \mathbb{R}^{d_{out}}$}
      	  \State {\bfseries Internal:} $\vec{\phi}\Leftarrow\vec{\phi}+s\cdot W_{i,\cdot}$
          \If{$c_j>0$}
          	  \State $\vec{\phi}\Leftarrow\vec{\phi}+\vec{v}$
          	  \While{$max(|\vec{\phi}|) > \tfrac12$}
                \State $j \leftarrow argmax(|\vec{\phi}|)$
                \State s = $sign(\phi_i)$ 
                \State $\phi_i\leftarrow\phi_i-s$
                \State $[W_{out}]_{\cdot, j} \leftarrow [W_{out}]_{\cdot, j} - \eta \cdot   s\cdot\vec{c_{in}}$ \Comment Update to $W_{out}$
                \State HiddenLayer.Backward($s\cdot [W_{out}]_{\cdot, j}$)
              \EndWhile
          \EndIf
      \EndProcedure

	\EndIndent

  \EndProcedure
  \end{algorithmic}
  \captionsetup{width=0.95\textwidth}
\end{algorithm}

\section{Network Diagram}

\begin{figure}[H]
\centering
\includegraphics[width=1\textwidth]{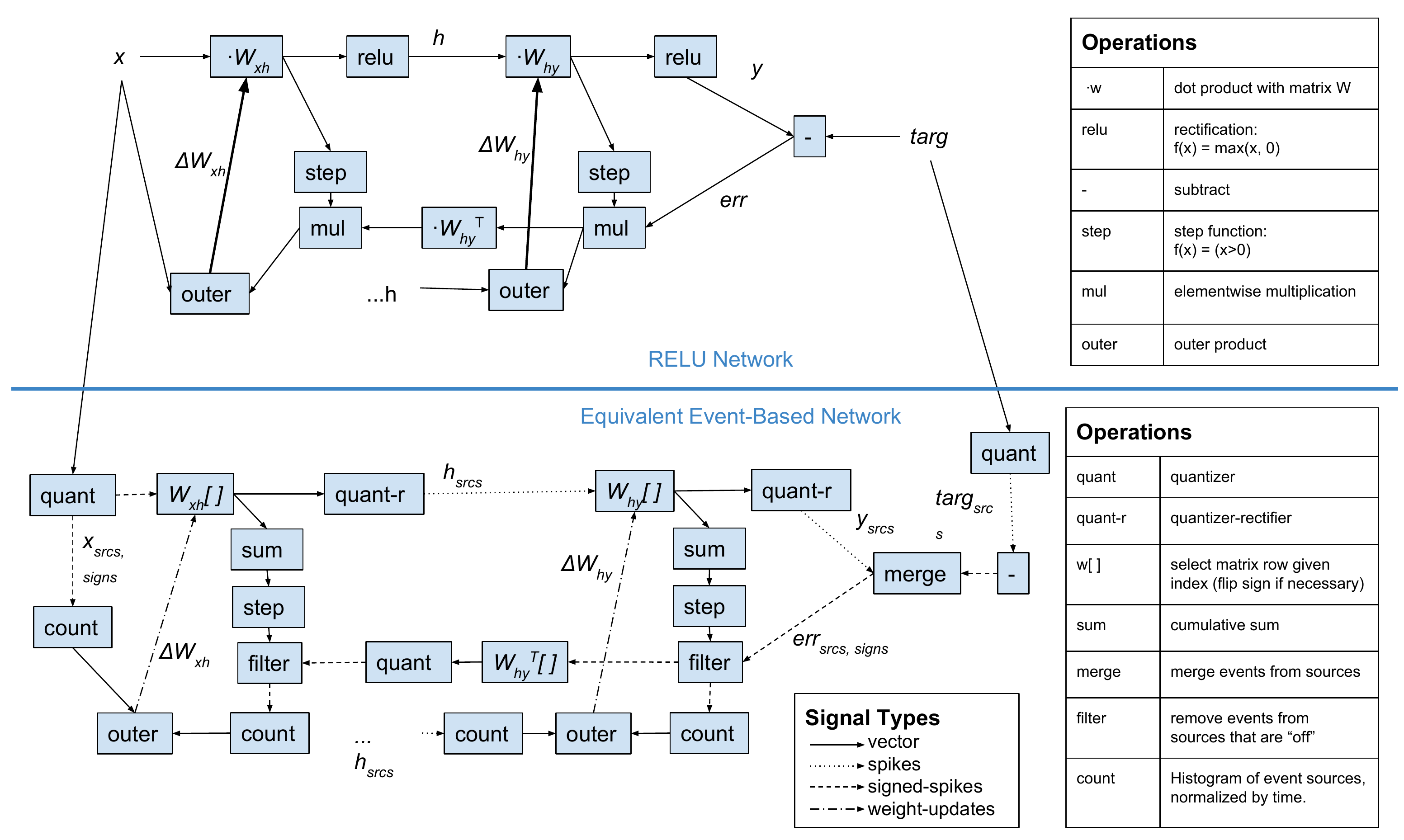}
\caption{\label{fig:spiking-mlp} The architecture of the Spiking MLP.  On the top, we have a conventional neural network of rectified linear units with one hidden layer.  On the bottom, we have the equivalent spiking network. 
}
\end{figure}

\section{Hyperparameters}
Our spiking architecture introduced a number of new hyperparameters and settings that are unfamiliar with those used to regular neural networks.  We chose to evaluate these empirically by modifying them one-by-one as compared to a baseline.  

\begin{itemize}[topsep=0pt,itemsep=-1ex,partopsep=1ex,parsep=1ex]
\item Fractional Updates.  
\begin{itemize}[topsep=0pt,itemsep=-1ex,partopsep=1ex,parsep=1ex]
\item False (Baseline): We use the standard stochastic-gradient descent method
\item True: We use our new Fractional Stochastic Gradient Descent method - described in section \ref{sec:weightupdates}  
\end{itemize}
\item Depth-First
\begin{itemize}[topsep=0pt,itemsep=-1ex,partopsep=1ex,parsep=1ex]
\item False (Baseline): Events are propagated "Breadth-first", meaning that, at a given time-step, all events are collected from the output of one module before any of their child-events are processed.
\item True: If an event from module A creates child-events from module B, those are processed immediately, before any more events from module A are processed.   
\end{itemize}
\item Smooth Weight Updates
\begin{itemize}[topsep=0pt,itemsep=-1ex,partopsep=1ex,parsep=1ex]
\item False (Baseline): The weight-update modules take in a count of spikes from the previous layer as their input.    
\item True: The weight-update modules take the rectified cumulative sum of the pre-quantized vectors from the previous layer - resulting in a smoother estimate of the input.  
\end{itemize}
\item Backwards-Quantization:
\begin{itemize}[topsep=0pt,itemsep=-1ex,partopsep=1ex,parsep=1ex]
\item No-Reset-Quantization (Baseline): The backwards quantization modules do not reset their $\vec{\phi}$s with each training iteration. 
\item Random: Each element of $\vec{\phi}$ is randomly selection from the interval $[-\frac12,\frac12]$ at the start of each training iteration. 
\item Zero-Reset: The backwards quantizers reset their $\vec{\phi}$s to zero at the start of each training iteration.    
\end{itemize}
\item Number of time-steps: How many time steps to run the training procedure for each sample (Baseline is 10).
\end{itemize}
Since none of these hyperparameters have obvious values, we tested them empirically with a network with layer sizes [784-200-200-10], trained on MNIST.  Table \ref{tab:hyperparams} shows the affects of these hyperparameters.

\begin{table}[H]
\centering
\caption{Percent error on MNIST for various settings.  We explore the effects of different network settings by changing one at a time, training on MNIST, and comparing the result to a baseline network.  The baseline network has layer-sizes [784, 200, 200, 10], uses regular (non-fractional) stochastic gradient descent, uses Breadth-First (as opposed to Depth-First) event ordering, does not use smooth weight updates, uses the "No-reset" scheme for its backward pass quantizers, and runs for 10 time-steps on each iteration.}
\label{tab:hyperparams}
\begin{tabular}{|l|l|}
\hline
\textbf{Variant}               & \textbf{\% Error} \\ \hline
 \hline\hline
Baseline                       &      3.38      \\ \hline
Fractional Updates             &      3.10      \\ \hline
Depth-First Propagation        &     81.47      \\ \hline
Smooth Gradients               &      2.85      \\ \hline
Smooth \& Fractional 		   &      3.07      \\ \hline
Back-Quantization = Zero-Reset &     87.87      \\ \hline
Back-Quantization = Random     &      3.15      \\ \hline
5 Time Steps                   &      4.41      \\ \hline
20 Time Steps                  &      2.65      \\ \hline
\end{tabular}
\end{table}

Most of the Hyperparameter settings appear to make a small difference.  A noteable exception is the Zero-Reset rule for our backwards-quantizing units - the network learns almost nothing throughout training.  The reason for this is that the initial weights, which were drawn from $\mathcal{N}(0, 0.01)$ are too small to allow any error-spikes to be sent back (the backward-pass quantizers never reach their firing thresholds).  As a result, the network fails to learn.  We found two ways to deal with this: ``Back-Quantization = Random'' initializes the $\vec{\phi}$ for the backwards quantizers randomly at the beginning of each round of training.  ``Back-Quantization = No-Reset'' simply does not reset $\vec{\phi}$ in between training iterations.  In both cases, the backwards pass quantizers always have some chance at sending a spike, and so the network is able to train.  It is also interesting that using Fractional Updates (FSGD) gives us a slight advantage over regular SGD (Baseline).  This is quite promising, because it means we have no need for multiplication in our network - As Section \ref{sec:weightupdates} explains, we simply add a column to the weight matrix every time an error spike arrives.  We also observe that using the rectified running sum of the pre-quantization vector from the previous layer as our input to the weight-update module (Smooth Gradients) gives us a slight advantage.  This is expected, because it is simply a less noisy version of the count of the input spikes that we would use otherwise.

\section{Event Routing}
\label{app:routing}

\begin{figure}[H]
\centering
\includegraphics[width=.5\textwidth]{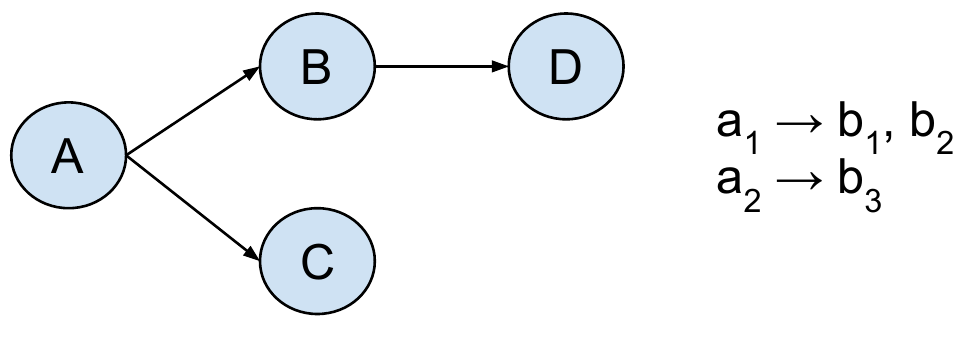}
\caption{\label{fig:simple-graph} A simple graph showing 4 modules.  Module A generates an event $a_1$ that causes two events: $b_1$ and $b_2$.  These are then distributed to downstream modules.  The order in which events are processed depends on the routing scheme.}
\end{figure}

Since each event can result in a variable number of downstream events, we have to think about the order in which we want to process these events.  There are two issues:

\begin{enumerate}
\item In situations where one event is sent to multiple modules, we need to ensure that it is being sent to its downstream modules in the right order.  In the case of the SMLP, we need to ensure that, for a given input, its child-events reach the filters in the backward pass before its other child-events make their way around and do the backward pass.  Otherwise we are not implementing backpropagation correctly.
\item  In situations where one event results in multiple child-events, we need to decide in which order to process these child events and their child events.   For this, there are two routing schemes that we can use: Breadth-first and depth-first.  We will outline those with the example shown in Figure \ref{fig:simple-graph}.  Here we have a module $A$ that responds to some input event by generating two events: $a_1$ and $a_2$.  Event $a_1$ is sent to module B and triggers events $b_1$ and $b_2$.  Event $a_2$ is sent and triggers event $b_3$.  Table \ref{tab:routers} shows how a breadth-first vs depth-first router will handle these events.  
\end{enumerate}

\begin{table}[!htb]
\begin{center}
 \begin{tabular}{||c c||} 
 \hline
 Breadth-First & Depth-First \\ [0.5ex] 
 \hline\hline
  $B(a_1)$ &  $B(a_1)$ \\ 
  $B(a_2)$ &  $D(b_1)$ \\ 
  $C(a_1)$ &  $D(b_2)$ \\ 
  $C(a_2)$ &  $C(a_1)$ \\ 
  $D(b_1)$ &  $B(a_2)$ \\ 
  $D(b_2)$ &  $D(b_3)$ \\ 
   $D(b_3)$ &  $C(a_2)$ \\ 
 \hline
\end{tabular}
\end{center}
 \caption{Depth-First and Breadth-First routing differ in their order of event processing.  This table shows the order of event processed in each scheme.  Refer to \ref{fig:simple-graph}.}
  \label{tab:routers}
\end{table}

Experimentally, we found that Breadth-First routing performed better on our MNIST task, but we should keep an open mind on both methods until we understand why.

\end{document}